\documentclass[11pt]{article}

\usepackage[preprint]{acl}

\usepackage{times}
\usepackage{latexsym}

\usepackage[T1]{fontenc}

\usepackage[utf8]{inputenc}
\usepackage{microtype}
\usepackage{inconsolata}
\usepackage{graphicx}
\usepackage{booktabs}       
\usepackage{microtype}      
\usepackage{url}            
\usepackage{amsfonts}       
\usepackage{nicefrac}       
\usepackage{amsmath}
\usepackage{bm, upgreek}
\usepackage{amssymb}
\usepackage{array}
\usepackage{multirow}
\usepackage{mathtools}
\usepackage{arydshln}
\usepackage{tabularx}
\usepackage{color}
\usepackage{soul}
\usepackage{caption}
\usepackage{comment}
\usepackage{adjustbox}
\usepackage{hyperref}
\usepackage{cleveref}
\usepackage[ruled,vlined]{algorithm2e}
\usepackage{makecell}
\usepackage{diagbox}
\usepackage{subfigure}
\usepackage{subcaption}
\usepackage{xcolor, soul}
\usepackage{wrapfig}
\usepackage{colortbl}
\usepackage[table]{xcolor}
\usepackage{pifont}
\usepackage[most]{tcolorbox}
\usepackage{listings}
\usepackage[frozencache]{minted}
\tcbuselibrary{listings, breakable}

\definecolor{bluecolor}{HTML}{0000FF}
\definecolor{greencolor}{HTML}{8CD0A4}
\definecolor{yellowcolor}{HTML}{F9D17C}
\definecolor{redcolor}{HTML}{FF0000}
\definecolor{Gray}{gray}{0.85}

\newcolumntype{a}{>{\columncolor{Gray}}c}

\definecolor{black}{rgb}{0,0,0}

\title{User-Oriented Multi-Turn Dialogue Generation with Tool Use at scale}

\author{Jungho Cho\thanks{Equal contribution} \\
  Upstage AI \\
  \texttt{christopher@upstage.ai} \\\And
  Minbyul Jeong\footnotemark[1] \\
  Upstage AI \\
  \texttt{minstar@upstage.ai} \\ \And
  Sungrae Park \\
  Upstage AI \\ 
  \texttt{sungrae.park@upstage.ai} \\
  }

\begin{document}
\maketitle

\begin{abstract}
The recent paradigm shift toward large reasoning models (LRMs) as autonomous agents has intensified the demand for sophisticated, multi-turn tool-use capabilities.
Yet, existing datasets and data-generation approaches are limited by static, predefined toolsets that cannot scale to the complexity of open-ended human-agent collaboration.
To address this, we initially developed a framework for automated task-oriented multi-turn dialogue generation at scale, utilizing an LRM-based simulator to dynamically generate high-value, domain-specific tools to solve specified tasks.

However, we observe that a purely task-oriented design often results in "solely task-solving" trajectories, where the agent completes the objective with minimal interaction, failing to generate the high turn-count conversations seen in realistic scenarios.
To bridge this gap, we shift toward a user-oriented simulation paradigm.
By decoupling task generation from a dedicated user simulator that mimics human behavioral rules—such as incremental request-making and turn-by-turn feedback—we facilitate more authentic, extended multi-turn dialogues that reflect the iterative nature of real-world problem solving.
Our generation pipeline operates as a versatile, plug-and-play module capable of initiating generation from any state, ensuring high scalability in producing extended tool-use data.
Furthermore, by facilitating multiple task completions within a single trajectory, it yields a high-density dataset that reflects the multifaceted demands of real-world human-agent interaction.
\end{abstract}

\section{Introduction}
\begin{figure}[t]
\centering
\includegraphics[width=.85\columnwidth]{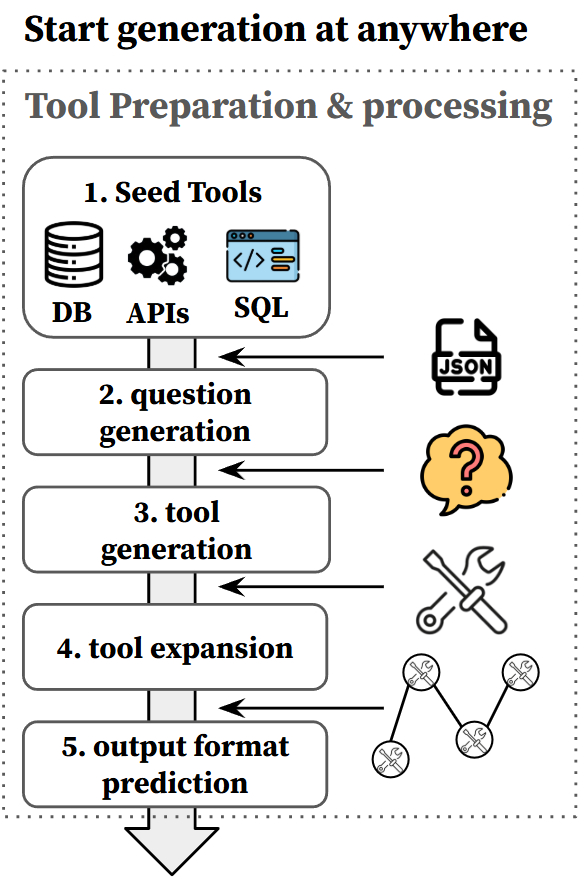}
\caption{
Plug-and-Play Tool Preparation Module. A modular pipeline for dynamic tool synthesis and preprocessing, designed to initiate multi-turn data generation from any arbitrary state.
}
\label{fig:motivation}
\vspace{-0.3cm}
\end{figure}
The evolution of large language models (LLMs) has reached a critical inflection point, transitioning from text-generative systems into large reasoning models (LRMs) acting as autonomous agents~\citep{guo2025deepseek, yang2025qwen3, yehudai2025survey}.
This change is driven by advances in both \textit{reasoning} and \textit{tool use}, grounded in a core set of agentic capabilities: the high-level decision-making and planning required to decompose complex tasks; the technical precision of tool choice and argument generation; the analytical rigor of result analysis and error handling; and the foundational memory and environment awareness needed to maintain context in dynamic settings~\citep{wang2023voyager, liu2023agentbench, mialon2023augmented, wu2024autogen, wang2024survey, xi2025rise}.
While tool use grounds reasoning in the real world, the ultimate objective of LRMs is to orchestrate these interdependent capabilities to support extended, multi-turn interactions that reflect the dynamics of real-world human–agent collaboration.

Despite the growing capabilities of LRMs, progress remains constrained by the lack of high-quality and diverse training data.
Most existing datasets rely on static, predefined toolsets, which inadequately capture the open-ended and evolving nature of real-world human–agent collaboration~\citep{team2025kimi, zhang2025nemotron, prabhakar2025apigen}.
Agents trained under such fixed schemas often struggle to generalize beyond seen domains or to reason over unfamiliar tool compositions.
Moreover, many data-generation pipelines implicitly favor single-shot trajectories\footnote{We define the trajectory as a sequence of API Calls, which are related (or correlated) with tool arguments for solving tasks}, in which a user poses a complex request and the agent responds with an optimal tool-use sequence in a single task.
While efficient, these interactions fail to reflect the iterative, incremental, and often noisy nature of real-world human–agent collaboration.

To overcome these limitations, we first developed an automated framework for large-scale task-oriented dialogue generation.
Leveraging an LRM-based simulator, the framework dynamically synthesizes domain-specific tools and database schemas (e.g., SQL-style read/write operations), along with corresponding tasks and evaluation rubrics.
While this successfully scaled the volume of data, we observed an efficiency trap: the simulator, acting as a perfect task-solver, tended to complete objectives with the minimum number of turns.
These solely task-solving trajectories lacked the back-and-forth dialogue—clarifications, incremental requests, and feedback loops—that define realistic human interaction.

To address this efficiency-driven bias, we propose a user-oriented simulation paradigm.
Our approach decouples the objective (the "Task") from the interaction (the "User").
By employing a dedicated user simulator governed by human behavioral rules—such as asking for only one subtask at a time and providing turn-by-turn feedback—we force the agent to navigate extended, multi-turn dialogues.

Our generation pipeline consists of three key components:
(1) Dynamic Tool \& Task Synthesis: Instead of relying on fixed APIs, our LRM-based generator creates unique, rubric-backed tasks grounded in synthesized database schemas, ensuring the agent learns to reason over diverse structures.
(2) Plug-and-Play Scalability: The generation pipeline is modular. It can initiate a simulation from any state—whether starting from a blank slate or injecting a tool-use requirement into an ongoing conversation—making it highly versatile for data augmentation.
(3) High-Density Trajectories: By allowing multiple task completions within a single conversation thread, we produce a "high-density" dataset. This reflects the multifaceted nature of real-world use cases, where a user might update a record, query a trend, and request a summary all within a single session.

Empirical results on agentic benchmarks, including BFCL~\citep{patilberkeley} and $\tau$2~\citep{barres2025tau}, demonstrate that models trained on our data achieve consistently stronger multi-turn performance and more reliable tool usage, particularly in long-horizon and stateful domains.
Moreover, consistency analysis under repeated executions shows that our models sustain correct tool-use behavior across multiple trials, rather than relying on isolated successes.
Our findings highlight the importance of user-oriented interaction modeling and execution-grounded supervision for training robust and realistic agentic reasoning models.
\section{Related Works}
\paragraph{Reasoning Models and Tool-use Benchmarks.}
Recent advancements in LRMs have catalyzed the development of benchmarks designed to evaluate autonomous agents in tool-mediated environments~\citep{guo2025deepseek, yang2025qwen3, team2025kimi, team2025tongyi, zeng2025glm}.
Early benchmarks primarily focused on single-turn tool invocation or static API selection within constrained domains~\citep{mialon2023gaia, qin2023toolllm, lee2025fhir}.
However, as the field shifted toward more complex problem solving, datasets like StableToolBench~\citep{guo2024stabletoolbench}, BFCL~\citep{patil2024gorilla}, and $\tau$ benchmarks~\citep{yao2024tau, barres2025tau} emerged to test the model's ability to navigate vast API landscapes~\citep{liu2025mcpeval, xu2025medagentgym, xi2025agentgym}.

\begin{figure*}[t]
\centering
\includegraphics[width=.9\textwidth]{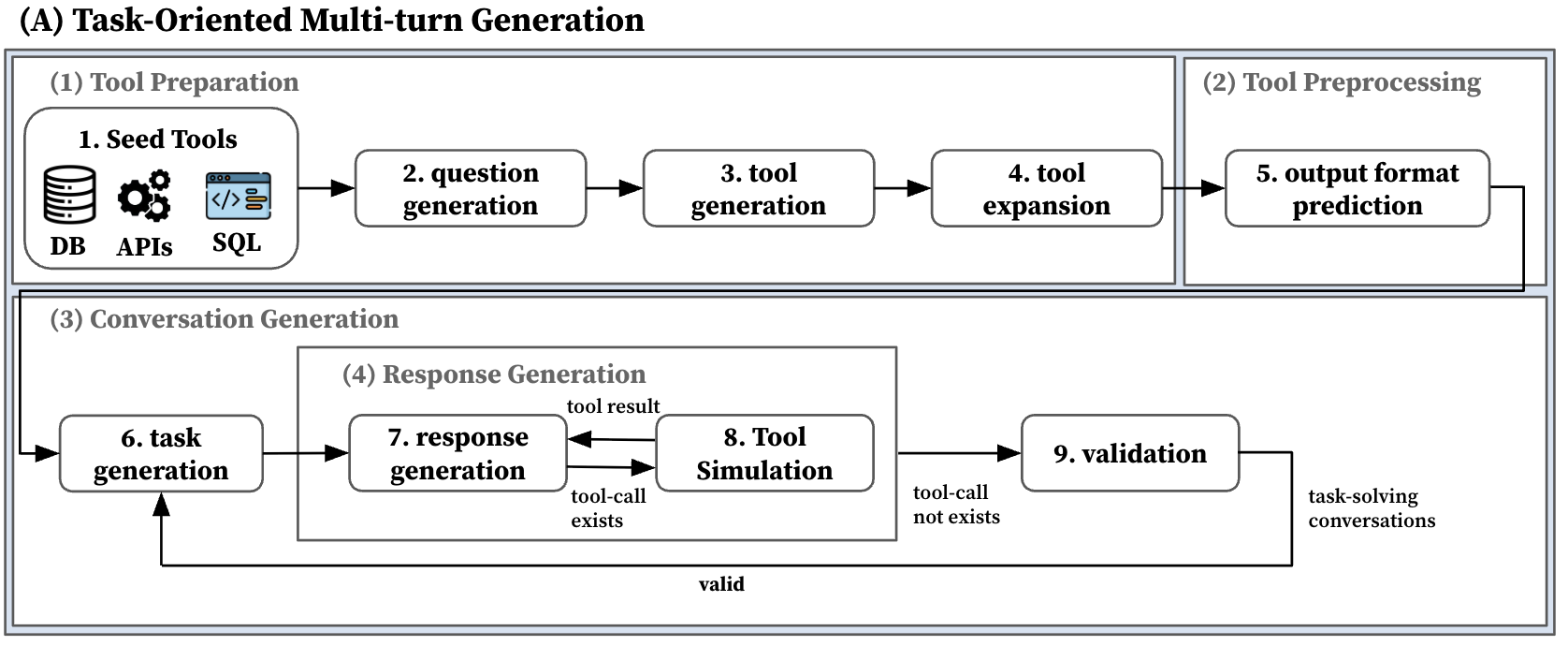}
\caption{
Task-Oriented Multi-Turn Conversation Generation Pipeline. An automated framework that generates tool-use trajectories focused on efficient task completion through direct simulator-based responses.
}
\label{fig:task_oriented}
\end{figure*}

\paragraph{Evolution of Tool-use Agents.}
Beyond benchmarking, the paradigm for tool-use agents has evolved from simple function-calling to sophisticated, autonomous orchestration.
While early frameworks enabled LLMs to parse queries and interpret results, they often relied on fixed toolsets, which inherently limited their adaptability to open-ended tasks~\citep{schick2023toolformer, hao2023toolkengpt}.
To address this, recent research has explored the dynamic creation of tools, such as generating reusable tools on the fly~\citep{cai2023large} or leveraging existing code repositories through ToolMaker~\citep{wolflein2025llm}.
Furthermore, specialized training strategies have been proposed to enhance agentic capabilities, including critique-informed planning~\citep{chen2025atlas}, fine-tuning on selective reasoning steps~\citep{yang2025lighthouse}, and decoupling reasoning from format following (e.g., Agent-FLAN)~\citep{chen2024agent}.
Despite these advances, most existing approaches still struggle to maintain long-term coherence in multi-turn interactions, a gap that our user-oriented simulation framework aims to bridge.

\paragraph{Synthetic Dialogue Generation for Agents.}
The scarcity of training data still requires high-quality synthetic data generation.
Despite these efforts, existing data generation approaches rely on fixed, predefined trajectories and toolsets (e.g., API graphs)~\citep{mitra2024agentinstruct, sengupta2024mag, arcadinho2024automated, tang2025synthesizing} and rigid schemas that fail to capture the stochastic and iterative nature of real-world dialogues, such as clarifying ambiguous user intents or handling incremental feedback~\citep{team2025kimi, prabhakar2025apigen, zhang2025nemotron}.
Consequently, there remains a significant gap in evaluating how reasoning models maintain coherence~\citep{barres2025tau} and adapt their tool-calling strategies over extended~\citep{zhang-etal-2024-probing}, multi-turn interactions—a limitation that underscores the need for a more dynamic, user-oriented simulation paradigm.

In our work, we decouple the generation process into independent stages, each with an individual component to be replaced easily with just modifying input-output format.
By architecting our pipeline as a versatile, plug-and-play module, we overcome the rigidity of previous approaches and enable the generation of high-density trajectories from any arbitrary state.
This allows for the synthesis of authentic, extended dialogues that incorporate incremental request-making and iterative feedback loops.
Consequently, our framework not only scales the production of domain-specific tools and database schemas dynamically but also ensures the generation of verifiable, multi-turn interactions that reflect the multifaceted and often noisy nature of real-world human-agent collaboration.
\section{Task-Oriented Multi-turn Generation}
To address the scarcity of high-quality agentic datasets, we developed a scalable, end-to-end pipeline designed to generate complex, multi-turn tool-use data.
While existing datasets like Nemotron~\citep{NemotronPostTrainingDatasetV1} provide a foundation with approximately 19K unique tools, they fall short of effective agentic training.
Our pipeline automates the entire lifecycle of data generation—from tool creation to task validation.
Designed as a plug-and-play module, the proposed framework allows for easy swapping of individual components by simply modifying input-output formats, ensuring diversity across domains and complexity in tool-interaction patterns.
An overview of the pipeline is illustrated in Figure~\ref{fig:task_oriented}.

\begin{figure*}[t]
\centering
\includegraphics[width=.9\textwidth]{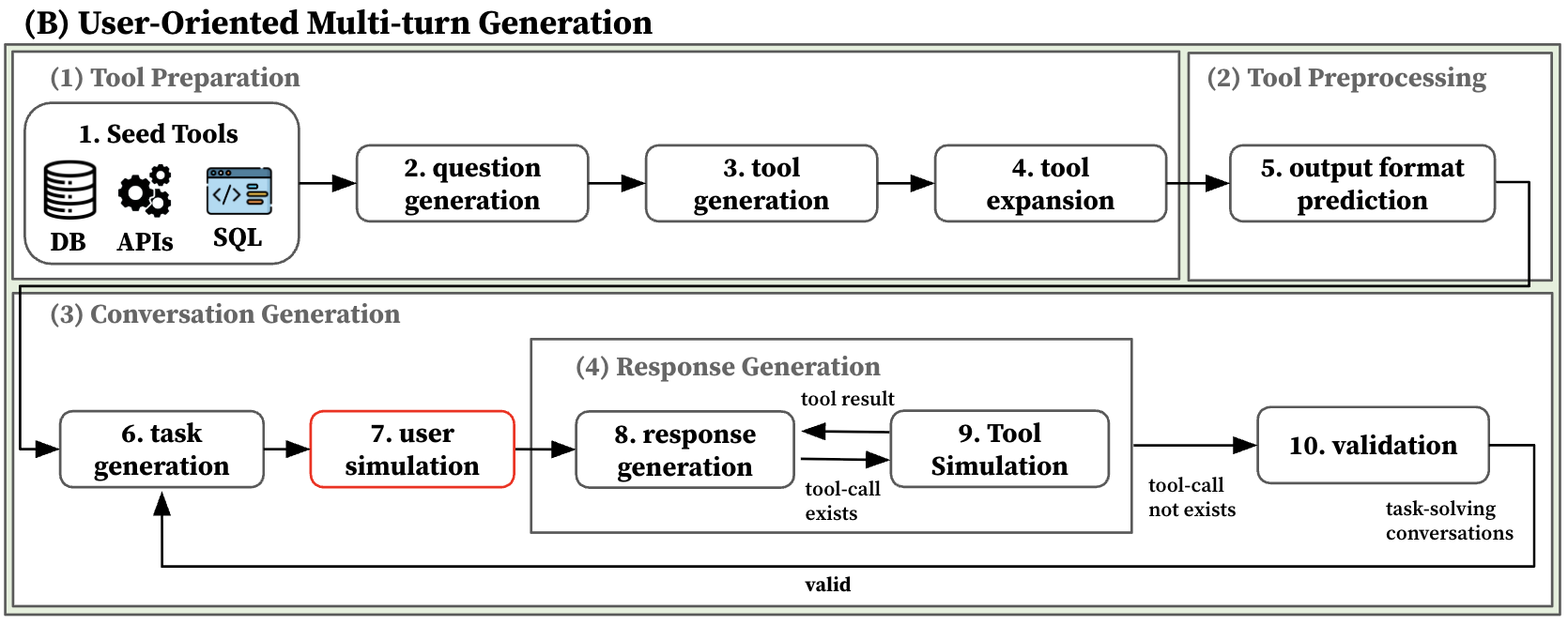}
\caption{
User-Oriented Multi-Turn Conversation Generation Pipeline. A framework that decouples tasks from interaction by employing a dedicated user simulator to mimic incremental human feedback and request-making.
}
\label{fig:user_oriented}
\end{figure*}

\subsection{Tool preparation}
The tool preparation stage aims to construct a diverse and realistic toolset from a minimal initial seed tool.
We begin by generating realistic user questions inspired by the existing seed tools, such as databases and APIs, to ensure that the synthesized tools are grounded in practical use cases.
Conditioned on these questions, the model generates detailed tool specifications, including tool names, natural language descriptions, and required parameters, such that each tool can programmatically solve the intended task.
To further expand domain coverage and interaction diversity, the framework analyzes the initial toolset to identify functional gaps and proposes up to ten complementary tools, resulting in a richer and more expressive toolset.
We detailed the prompt in Appendix~\ref{app:prompt}.

\subsection{Tool preprocessing}
Before conversation simulation, we perform tool preprocessing to ensure structural and semantic consistency across the generated toolset.
Specifically, the model is instructed to predict a JSON schema for the return value of each tool, making tool outputs explicit and machine-verifiable.
Schema definitions are generated in a multi-turn conversational manner, allowing the model to reason over previously defined tools and maintain input-output consistency.
As a result, shared entities such as \texttt{user\_id} or \texttt{timestamps} preserve consistent data types and semantics across different tools and interaction turns.

\subsection{Conversation Generation}
\paragraph{Task Generation.}
Given the preprocessed toolset, we generate multi-turn conversations by constructing structured, rubric-based tasks.
Following Kimi-K2~\citep{team2025kimi}, each task is categorized by difficulty level (easy, medium, or hard) and is accompanied by a detailed rubric that specifies success criteria, expected tool-use patterns (with placeholders for dynamic arguments), and intermediate evaluation checkpoints.
These components enable objective, step-level verification of agent behavior while encouraging complex reasoning and multi-step tool interactions.

\paragraph{Response Generation.}
In this stage, the generation model (here we use GPT-OSS-120b) produces responses that are validated for correctness and quality.
Since all tools are synthetic, an LRM-based simulator is employed to generate tool execution results conditioned on the provided arguments and the evolving conversation context.
To ensure temporal realism, the simulator maintains a randomized reference time while prioritizing user-specified temporal information when present.

\paragraph{Validation.}
Finally, a dedicated validation module compares the agent’s responses against the predefined rubrics, filtering trajectories based on semantic correctness and required tool invocations.
Only successful, high-density interaction trajectories are retained in the final dataset.
Despite the effectiveness of this task-oriented pipeline, we observe that it produces trajectories focused solely on efficient task completion with minimal interaction.
This limitation motivates the transition toward a user-oriented simulation paradigm, which we introduce in the following section.

\section{User-Oriented Multi-turn Generation}
Although the task-oriented pipeline effectively scales data volume, we observe that it frequently falls into an \emph{efficiency trap}, where a highly capable simulator completes complex objectives in a single turn with minimal interaction (see statistics in Table~\ref{tab:statistics}).
Such behavior produces trajectories that are optimized for task completion, failing to capture the incremental, exploratory, and iterative nature of realistic human–agent collaboration.
In addition, many simulator-based approaches depend on synthetic tool outputs, which limits the faithfulness and verifiability of the resulting interactions.

To better reflect realistic usage patterns, we introduce a user-oriented multi-turn generation paradigm that explicitly models user behavior and interaction dynamics (see Figure~\ref{fig:user_oriented}).
Building on the same tool-use abstraction introduced in earlier sections, this paradigm further grounds tool interactions in executable environments, enabling multi-turn trajectories whose intermediate states and outcomes are consistently maintained across turns rather than being implicitly assumed from a single-shot request.

\begin{table}[t]
\centering
{\resizebox{\columnwidth}{!}{
\begin{tabular}{lcccc}
\toprule
\textbf{Source data} & \textbf{Step} & \textbf{Turn} & \textbf{Task} & \textbf{Samples} \\
\midrule
\multicolumn{5}{l}{\textit{\textbf{Task-oriented}}} \\
\textbf{Nemotron}    & 3.95 (89) & 12.84 (178) & 1.63 (18) & 161,608 \\ 
\midrule
\multicolumn{5}{l}{\textit{\textbf{User-oriented}}} \\
\textbf{Nemotron}    & 3.45 (121) & 21.79 (596) & 2.48 (20) & 177,375  \\ 
\textbf{Tau2}        & 3.05 (63) & 36.16 (780) & 3.02 (20) & 4,138 \\ 
\midrule
\multicolumn{5}{l}{\textit{\textbf{User-oriented + Tool-Execution}}} \\
\textbf{Tau2}        &  2.43 (23) & 17.15 (294) & 1.6 (10) & 2,174 \\ 
\textbf{SQL}         & 3.74 (322) & 30.85 (680) & 1.86 (11) & 16,618   \\ \bottomrule
\end{tabular}}}{}
\caption{Statistics of Generated Datasets.
Comparison of conversation density (steps and turns) across task-oriented and user-oriented paradigms; values in parentheses represent the maximum observed counts. We define each column in Appendix~\ref{app:definition}}
\label{tab:statistics}
\vspace{-0.3cm}
\end{table}


\paragraph{Descriptive Task Generation.}
To support user-oriented interaction, we modify the generation pipeline to produce descriptive tasks instead of direct user questions.
Rather than emitting a fully specified natural-language query, the model first generates a declarative statement that describes the user’s ultimate objective in a self-contained manner.
These descriptive tasks serve as latent goals that guide the user simulator, which must then realize them incrementally through multi-turn interaction.

For settings that involve structured tools, the task generator is additionally conditioned on concrete environment information, such as database schemas and limited data views.
This grounding ensures that generated tasks remain feasible with respect to actual table structures and supported operations, including reading, updating, and combining records.
To avoid introducing unsupported assumptions—particularly when only partial information is available—the generator is encouraged to phrase uncertain facts as information that must be retrieved or confirmed through subsequent tool use, rather than treating them as known in advance.

For longer trajectories, additional descriptive tasks are introduced either as natural extensions of previous tool outcomes or as independent but contextually coherent objectives within the same domain.
Throughout this process, task complexity is explicitly controlled (easy, medium, or hard), ensuring that the resulting dialogues remain appropriately challenging while remaining consistent with earlier task-oriented settings.

\begin{figure}[t]
\centering
\includegraphics[width=\columnwidth]{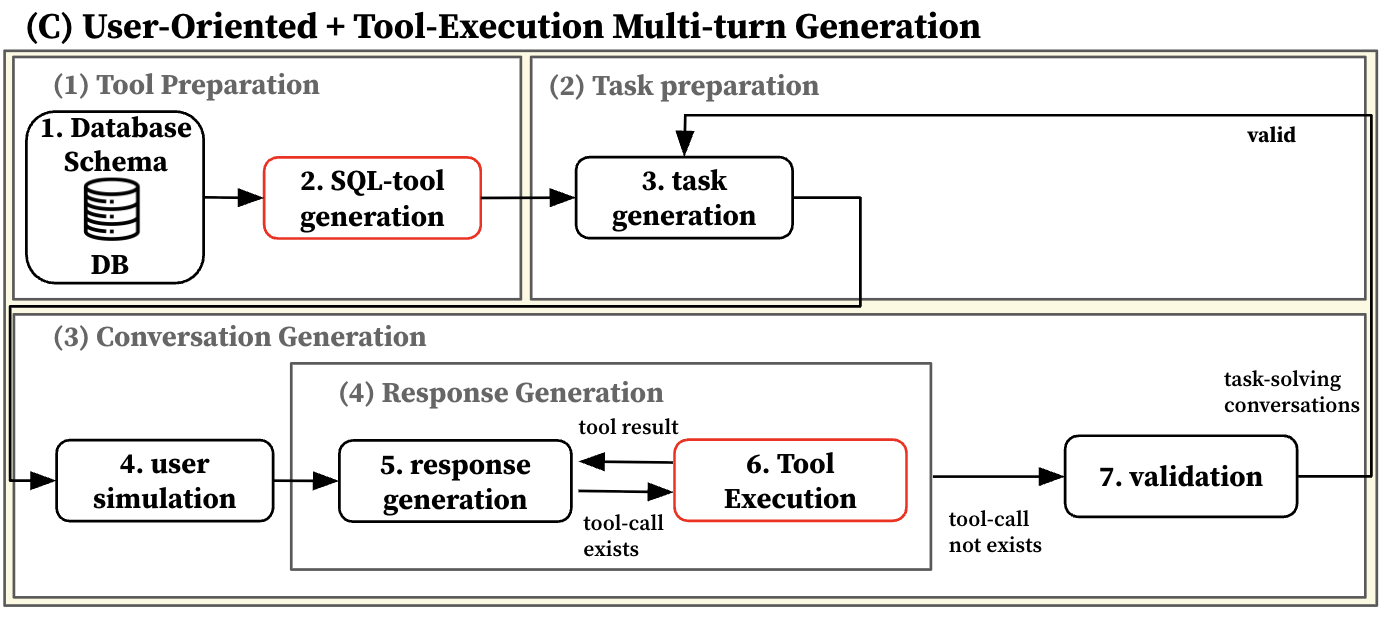}
\caption{
User-oriented Tool-Execution Multi-turn Conversation Generation Pipeline. This pipeline integrates a SQL-tool generation module grounded in real-world database schemas with a dedicated user simulator to produce verifiable, high-fidelity multi-turn dialogues.
}
\label{fig:user_oriented_execution}
\vspace{-0.3cm}
\end{figure}

\begin{table*}[t]
\centering
{\resizebox{1.0\textwidth}{!}{
\begin{tabular}{l ccccc ccc}
\toprule
\multicolumn{1}{c}{\multirow{2}{*}{\textbf{Model}}} & \multicolumn{5}{c}{\textbf{BFCL}} & \multicolumn{3}{c}{\textbf{$\tau$2}}  \\ \cmidrule(lr){2-6} \cmidrule(lr){7-9}
\multicolumn{1}{c}{}  & \multicolumn{1}{c}{\textbf{Multi-turn}} & \multicolumn{1}{c}{\textbf{Live}} & \multicolumn{1}{c}{\textbf{Non-Live}} & \multicolumn{1}{c}{\textbf{Hall. (Rel)}} & \multicolumn{1}{c}{\textbf{Hall. (Irrel)}} & \multicolumn{1}{c}{\textbf{Airline}} & \multicolumn{1}{c}{\textbf{Retail}} & \multicolumn{1}{c}{\textbf{Telecom}}\\ \midrule
\multicolumn{9}{l}{\textit{\textbf{Proprietary Models}}} \\
Claude Sonnet 4.5 (fc)$\dagger$        &   60.9    &   81.1   &   88.6  &  68.8 & 86.3     &  70.0  &  86.2 & 98.0              \\ 
GPT-5.1-mini$\dagger$            &   27.4    &   58.6  &   70.2   &  68.8 & 91.8    & -  & - & -                \\ 
GPT-5.1$\dagger$                  &   36.1    &   59.0   &  73.0   & 68.8 & 91.4    & 77.9 & - & 95.6          \\
Gemini-2.5-Pro$\dagger$  & 29.3 & 63.8 & 85.6 & 43.8 & 91.5 & - & - & - \\
Gemini-3.0-Pro$\dagger$  & - & -  & - & - & - & 73.0 & 85.3 & 85.4 \\ \midrule
\multicolumn{9}{l}{\textit{\textbf{Open-Sourced Models}}} \\
xLAM-2-3b-fc-r$\dagger$           &   56.0    &   58.7   &   82.9  &   94.4   & 57.9 &  32.0  &  44.4 & -                \\ 
xLAM-2-8b-fc-r$\dagger$           &   69.3   &   66.7   &   84.4  &   83.3   & 64.1 &  35.2  &  58.2 & -               \\ 
xLAM-2-32b-fc-r$\dagger$          &   66.4    &   73.8   &   89.5  &   83.3 &  76.3  &  45.0  &  64.3 & -                \\
Qwen3-4B-Thinking-2507   &   48.1 &  82.9 &  86.3 &   100.0 &  78.9 &  46.0 &  56.1 & 21.1  \\ 
Qwen3-30B-A3B-Thinking-2507                 &   53.8 &  84.1 &  89.6 &   100.0 &  80.6 &  56.0 &  54.4 & 22.8  \\ 
GPT-OSS-120b  &   51.3 &  72.6 &  37.5 &  75.0 &  85.5 &  56.4  & 75.3 & 59.5  \\
\midrule
\multicolumn{9}{l}{\textit{\textbf{Baselines}}} \\
Qwen3-4B-Thinking-2507 + \textsc{Apigen}  &   50.9 &  83.1 &  87.5 &   83.3 &  82.3 &  50.0  & 58.8 & 30.7               \\ 
Qwen3-30B-A3B-Thinking-2507  + \textsc{Apigen}   &   53.8 &  83.5 &  90.1 &   83.3 &  82.3 &  56.0  & 60.5 & 33.3               \\ 
Qwen3-4B-Thinking-2507 + \textsc{Nemotron}  &   52.1 &  85.8 &  88.2 &   75.0 &  85.5 &  44.0  & 50.9 & 26.3            \\ 
Qwen3-30B-A3B-Thinking-2507  + \textsc{Nemotron}  &   46.0 &  50.9  &  88.7 &   88.9 &  76.3 &  54.0  & 48.2 & 28.1            \\ 
\midrule
\multicolumn{9}{l}{\textit{\textsc{\textbf{Ours}}}} \\
Qwen3-4B-Thinking-2507 + \textsc{Ours}  &   52.7 &  84.9 &  89.6 &   83.3 &  80.6 &  52.0  & 57.0 & 36.8          \\ 
Qwen3-30B-A3B-Thinking-2507  + \textsc{Ours}   &   55.5 &  86.5 &  90.1 &   88.9 &  83.8 &  56.0  & 57.8 & 42.1           \\ 
\bottomrule
\end{tabular}}}
\caption{Agentic benchmark results across proprietary and open-source models. $\dagger$ refers to the reported scores.
The table compares performance on the Berkeley Function Calling Leaderboard (BFCL)~\citep{patilberkeley} and $\tau2$~\citep{barres2025tau} benchmarks, highlighting the improvements gained from our data generation pipeline compared to baselines like \textsc{Apigen}~\citep{prabhakar2025apigen} and \textsc{Nemotron}~\citep{NemotronPostTrainingDatasetV1}.}
\label{tab:main_table}
\vspace{-0.3cm}
\end{table*}

\paragraph{User Simulation Interaction Loop.}
At the core of the proposed paradigm is a dedicated user simulator governed by simple yet expressive behavioral rules.
Given a descriptive task, the simulator identifies the required sub-tasks but deliberately issues requests in a piecemeal fashion, typically asking for only one or two subtasks per turn.
This design encourages the assistant to engage in intermediate reasoning, clarification, and verification, rather than converging immediately on a final answer.

Unlike simulated pipelines, the simulator conditions its behavior on tool outputs that are produced through actual execution.
In practice, tools correspond to concrete operations such as parameterized database queries that are executed against a controlled environment, and their results are returned verbatim to the dialogue.
The simulator maintains contextual awareness by reviewing the assistant’s prior responses and tool outcomes to assess which components of the overall goal have been satisfied.
Based on this assessment, it provides turn-by-turn feedback, requests clarifications, or introduces follow-up questions until the objective is fully achieved.
A conversation is considered complete only when the simulator explicitly signals task completion by setting \texttt{is\_task\_complete} to true.

\paragraph{High-Density Multi-turn Trajectories.}
By allowing multiple descriptive tasks to be addressed within a single conversational thread, the proposed pipeline naturally produces high-density multi-turn trajectories.
This setting mirrors realistic usage scenarios in which users perform a sequence of related actions—such as querying information, updating records, and requesting summaries—within a single session.
To maintain coherence, state changes introduced by tool use persist across turns within the same trajectory, while remaining isolated across different generation instances.

Moreover, the modular and plug-and-play design of the pipeline enables generation to begin from arbitrary intermediate states, significantly improving scalability and diversity for extended tool-use data.
Overall, the user-oriented paradigm complements the task-oriented pipeline by emphasizing interaction richness, temporal continuity, and verifiable tool use, which together are essential for training robust agentic reasoning models.

\paragraph{From Simulated Tools to Executable SQL-driven Agents.}
To overcome the scalability limits of static toolsets and the hallucination risks inherent in model-based simulations, we introduce a framework that synthesizes executable tool interfaces grounded in real-world relational databases (see Figure~\ref{fig:user_oriented_execution}).
By leveraging diverse schemata from open-source datasets like Spider~\citep{yu2018spider}, our pipeline automatically generates domain-specific functions mapped to complex SQL queries.
We visualized the domains and examples used in our generated data in Appendix~\ref{app:visualization} and~\ref{app:sql_example}.
This approach allows the agent to interact with a functional database engine in real-time during the dialogue generation process, ensuring that the tool outputs used for training are computationally verified and factually accurate.
Consequently, this SQL-backed synthesis enables the production of high-fidelity, multi-turn trajectories at scale, transforming the data generation pipeline from a closed-loop simulation into a verifiable, agentic execution environment.

\section{Experiments}

\subsection{Experimental Setups}
\paragraph{Training \& Inference.}
We perform full fine-tuning of reasoning models.
To balance the training data provided in Table~\ref{tab:statistics}, we downsampled the synthetic data generated from the Nemotron dataset due to the lower performance trend.
All models are trained for five epochs, and the checkpoint corresponding to the best validation performance is selected for final evaluation.
For inference, we serve the fine-tuned models using vLLM~\citep{kwon2023efficient}, enabling efficient long-context decoding and high-throughput evaluation.
Our experiments focus on two Qwen-family reasoning models~\citep{yang2025qwen3} with varying scales and generation model GPT-OSS-120b~\citep{agarwal2025gpt}.
This setup allows us to analyze the impact of model capacity on multi-turn, tool-augmented agent behavior.
We detailed the rest of the descriptions in Appendix~\ref{app:training_inference_details}.

\paragraph{Evaluation.}
To evaluate an agent’s robustness to noisy, incremental user requests and its ability to sustain coherent multi-turn tool interactions, we adopt two complementary agentic benchmarks: $\tau2$~\citep{barres2025tau} and the Berkeley Function Calling Leaderboard (BFCL)~\citep{patilberkeley}.
We detailed the evaluation sets in Appendix~\ref{app:evaluation}.

\paragraph{Source Data and Statistics.}
As summarized in Table~\ref{tab:statistics}, we construct a high-density tool-use dataset by leveraging seed tools from the \textsc{Nemotron}~\citep{NemotronPostTrainingDatasetV1} post-training dataset and structured tasks from the $\tau2$ database benchmark.
These seeds are expanded via our pipeline into diverse, domain-specific toolsets and executable database schemas.
Unlike task-oriented datasets, our generated trajectories frequently contain multiple task completions within a multi-turn conversation.
This design reflects realistic user sessions and enables more faithful training and evaluation of long-horizon agentic behavior.

\begin{table}[t]
\centering
{\resizebox{\columnwidth}{!}{
\begin{tabular}{l c c}
\toprule
\multicolumn{1}{c}{\multirow{2}{*}{\textbf{Generation Pipeline}}} & \multicolumn{1}{c}{\textbf{BFCL}} & \multicolumn{1}{c}{\textbf{Tau2}}  \\ \cmidrule{2-3} 
\multicolumn{1}{c}{}   & \multicolumn{1}{c}{\textbf{Multi-turn}} & \multicolumn{1}{c}{\textbf{Telecom}}\\ \midrule
\multicolumn{3}{l}{\textit{Task-oriented}} \\
Qwen3-4B-Thinking-2507 &   50.9 &  24.5           \\ 
Qwen3-30B-Thinking-2507  &   53.8 &  26.3        \\ 
\midrule
\multicolumn{3}{l}{\textit{User-oriented}} \\ 
Qwen3-4B-Thinking-2507  &   51.8 &  30.7           \\ 
Qwen3-30B-Thinking-2507   &   54.5 &  34.2        \\ 
\midrule
\multicolumn{3}{l}{\textit{User-oriented + Tool Execution}} \\
Qwen3-4B-Thinking-2507  &   52.7 &  35.1     \\ 
Qwen3-30B-Thinking-2507  &   54.9 &  40.4    \\ 
\bottomrule
\end{tabular}}}
\caption{Ablation study of the generation pipeline.}
\label{tab:ablation_results}
\vspace{-0.3cm}
\end{table}

\subsection{Experimental Results}
Table~\ref{tab:main_table} summarizes the agentic benchmark performance of models fine-tuned with different data generation pipelines.
Overall, models trained on our user-oriented synthetic data consistently outperform counterparts trained on prior baselines, including \textsc{Apigen}~\citep{prabhakar2025apigen} and \textsc{Nemotron}~\citep{NemotronPostTrainingDatasetV1} across both BFCL and $\tau$2 benchmarks.

Across model scales, the gains are most pronounced on $\tau$2, which explicitly evaluates robustness to incremental user requests and long-horizon interaction.
For both Qwen3-4B and Qwen3-30B, fine-tuning with our data (we refer to it as \textsc{Ours}) yields steady improvements over baseline datasets training, indicating that richer, multi-turn supervision improves the model’s ability to track goals, maintain state, and adapt tool-calling strategies over extended dialogues.
Models trained on \textsc{Nemotron} alone exhibit weaker performance on several $\tau$2 domains, suggesting that without using in-domain database is insufficient for capturing realistic user–agent interaction dynamics.

In Table~\ref{tab:ablation_results}, we further observe that grounding tool execution in real, executable environments plays a critical role.
Compared to purely simulated pipelines, our user-oriented + tool-execution setting produces the strongest overall results, particularly in Telecom domain.
The Telecom domain requires persistent state tracking and iterative refinement of user intent, and the improvements suggest that exposure to verifiable database-backed tool outputs encourages more faithful tool selection and error recovery behaviors.
Importantly, these gains do not come at the expense of function-calling accuracy: BFCL scores remain stable or improve slightly, indicating that increased conversational complexity does not degrade low-level tool invocation fidelity.
Taken together, the results demonstrate that user-oriented simulation with execution-grounded supervision jointly contribute to stronger agentic performance, especially in benchmarks that emphasize multi-turn coherence and interaction realism.


\begin{figure*}[t]
\centering
\includegraphics[width=\textwidth]{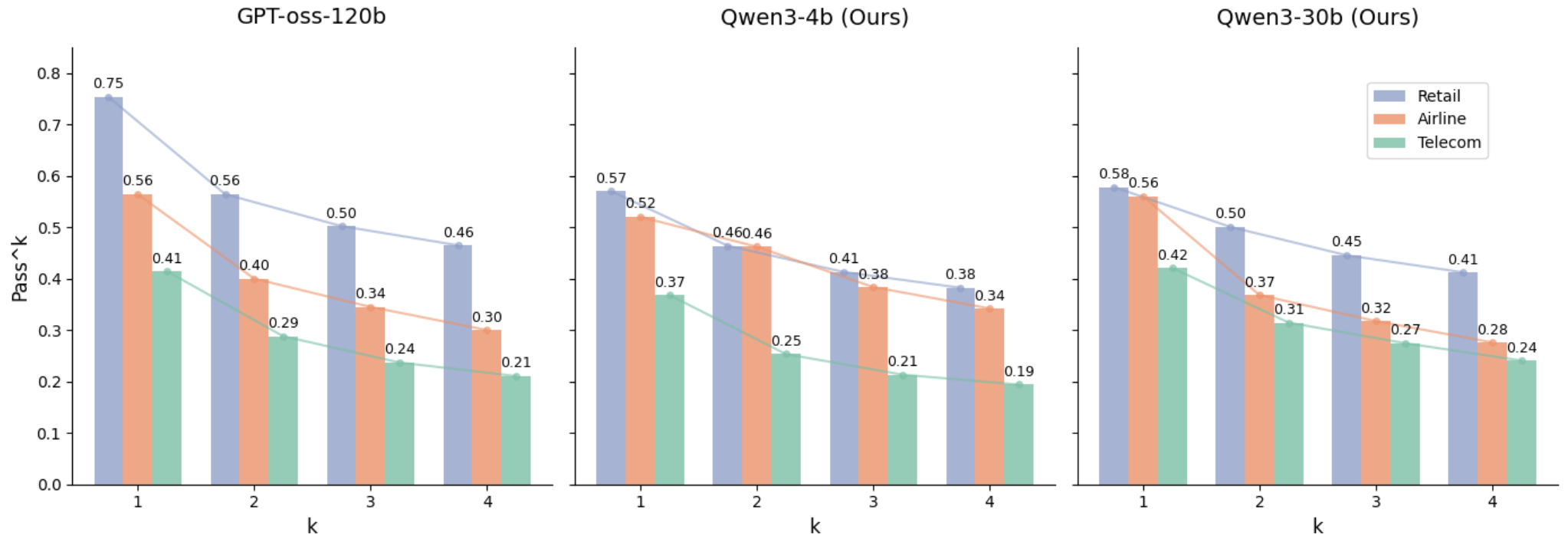}
\caption{
Consistency analysis across varying $k$ values. The charts illustrate the Pass$\string^$k performance for different models (GPT-OSS-120b, Qwen3-4b, and Qwen3-30b) across the Retail, Airline, and Telecom domains, showing how performance scales preserve in overall domains while increased $k$ values.
}
\label{fig:consistency}
\vspace{-0.3cm}
\end{figure*}
\begin{table}[t]
\centering
{\resizebox{\columnwidth}{!}{
\begin{tabular}{lccc}
\toprule
\textbf{Pipelines} & \textbf{Latency} & \textbf{Throughput} & \textbf{GPUs} \\
\midrule
\textit{\textbf{Task-oriented}} & 0.64 & 8,819 & 32 \\
\textit{\textbf{User-oriented}} & 4.11 & 4,079 & 32 \\
\bottomrule
\end{tabular}}}{}
\caption{Generation efficiency comparison between task-oriented and user-oriented pipelines.}
\label{tab:generation_efficiency}
\vspace{-0.3cm}
\end{table}
\section{Analysis}
\subsection{Generation Efficiency}
\label{app:generation_efficiency}
Table~\ref{tab:generation_efficiency} compares the inference efficiency of the task-oriented and user-oriented generation pipelines.
Experiments were conducted on NVIDIA H100 GPUs using the GPT-OSS-120B model with a tensor parallel size of 2.
The deployment consists of 4 nodes (32 GPUs in total), hosting 16 parallel model instances.
\textbf{Latency} is measured as the average wall-clock time in seconds per generated sample, while \textbf{Throughput} denotes the number of generated tokens per second aggregated across all GPUs. 
The user-oriented pipeline exhibits higher latency and lower throughput due to longer multi-turn interactions and increased generation complexity, reflecting a trade-off between interaction realism and generation efficiency.

\subsection{Consistency of Tool Usage}
To assess the consistency of tool usage rather than one-off success, following analysis of $\tau2$~\citep{barres2025tau}, we analyze model performance using the Pass$\string^$k metric, illustrated in Figure~\ref{fig:consistency}.
Concretely, the metric measures how often a model can correctly complete the same task when it is attempted repeatedly under identical conditions.

Models trained with our pipeline exhibit consistently higher Pass$\string^$k values across domains, indicating that correct tool usage is sustained across multiple trials.
This trend holds across the domains, suggesting that the gains are not domain-specific artifacts but reflect a general improvement in reliable tool execution.
Importantly, the preserve of performances across $k$ values is most pronounced in domains with higher interaction complexity and statefulness, such as Telecom.
In these settings, repeated correct execution requires not only accurate tool invocation but also robust tracking of intermediate states and user intent across turns.
The higher Pass$\string^$k scores therefore indicate that our training data encourages models to internalize stable tool-use strategies that generalize across repeated attempts.
Overall, by explicitly evaluating multiple trials per task, the Pass$\string^$k analysis confirms that our approach improves the consistency and robustness of tool usage, aligning with the goal of training agents that behave reliably under repeated, real-world usage rather than optimizing for isolated successes.


\section{Conclusion and Discussion}
In this work, we present a scalable, user-oriented simulation framework for multi-turn dialogue generation.
By architecturing our pipeline as a plug-and-play module, we overcome the rigidity of previous static approaches and enable the generation of high-fidelity and high-density trajectories.
This ensures the production of verifiable interactions that reflect the multifaceted and iterative nature of realistic user-agent communication.
Across our experiments, the results suggest that a user-oriented generation pipeline with tool execution plays a central role in improving long-horizon agent performance, as evidenced by substantial gains on $\tau2$ and its Telecom domain.
The transition from simulated tools to executable tools further highlights an important discussion point: execution-grounded supervision appears to encourage faithful tool selection, state tracking, and recovery behavior, particularly in environments where actions modify persistent state.

However, this increased realism introduces new challenges, including higher generation cost, tighter coupling between environment consistency and data quality, and increased brittleness under partial database visibility.
The SQL-based executable pipeline represents a promising direction toward scalability, demonstrating that realistic, stateful tool use can be extended beyond handcrafted benchmarks, although its impact on cross-domain generalization remains an open question.


\section*{Limitations}
While our user-oriented pipeline produces high-fidelity trajectories, it introduces higher computational costs and latency compared to the task-oriented pipeline (see Section~\ref{app:generation_efficiency}), as it requires multiple rounds of interaction between the simulator and the environment.
Unlike a task-oriented pipeline that often falls into an `efficiency trap' by completing objectives in a single turn, a user-oriented pipeline requires multiple rounds of iterative reasoning and interaction between the user simulator, the agent, and the execution environment.
This multi-turn exchange, while necessary for capturing the incremental nature of human collaboration, results in a higher cumulative token consumption and extended processing time per successful data sample.
Consequently, scaling this pipeline to millions of trajectories poses a practical challenge in terms of the total GPU hours and API costs required compared to more direct, single-shot data synthesis methods.

The transition to execution-grounded data generation introduces a tighter coupling between the consistency of the simulation environment and the overall quality of the resulting dataset.
Our SQL-backed tool-execution pipeline relies on a precise alignment between synthesized database schemas and the agent’s tool-calling logic; any discrepancy in state tracking across long-horizon interactions can lead to error propagation.
Furthermore, the model exhibits a certain degree of brittleness when presented with partial or ambiguous database views, occasionally struggling to maintain factual accuracy when the required information is not explicitly provided in the initial context.
These challenges underscore the need for more robust state-recovery mechanisms and sophisticated error-handling strategies within the simulation loop to ensure long-term trajectory coherence.

\section*{Acknowledgments}


\bibliography{custom}

\appendix
\label{sec:appendix}


\section{Prompt of User-oriented Multi-turn Conversation}
\label{app:prompt}

We wrote the overall prompt of our generation pipeline in the end of the manuscript.

\section{Step, Turn, and Task definition of Generated dataset}
\label{app:definition}
In this section, we provide the formal definitions (used in Table~\ref{tab:statistics}) and statistical breakdowns of the generated trajectories in our dataset.
To ensure high-fidelity simulation of human-agent collaboration, we categorize the complexity of our data using three primary metrics: Turns, Steps, and Tasks.

\begin{itemize}
\item Turn: A total number of discrete exchanges within a single session. This includes all User utterances, Assistant responses, and Tool invocations/outputs. A higher turn count typically indicates a more conversational and interactive session rather than a simple "one-shot" query.
\item Step: We defined it as the number of sequential tool-use iterations required to satisfy a single user request. For instance, if a user asks for a flight recommendation, the agent might.
\item Task: The number of high-level objectives assigned to the agent within a single session. It represents a complete functional goal (e.g., "Schedule a meeting" or "Analyze a financial report"). Multi-task trajectories test the agent's ability to maintain context across shifting goals.
\end{itemize}

\section{Domain Visualization of SQL-based Tool-execution Data}
\label{app:visualization}
\begin{figure}[h]
\centering
\includegraphics[width=\columnwidth]{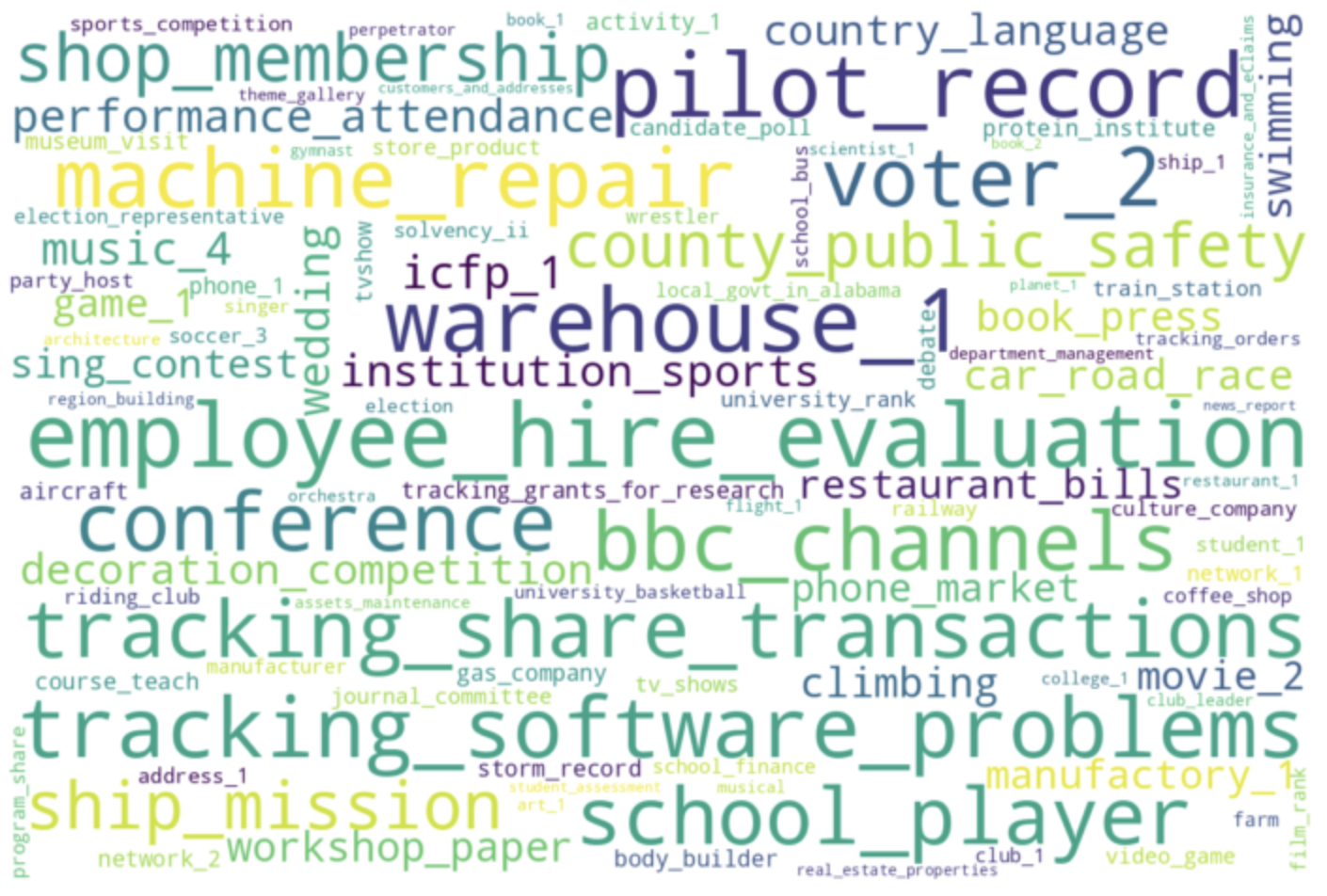}
\caption{
Domain Visualization of SQL-based User-Oriented Tool Execution Data. A word cloud visualizing the diverse, real-world domains synthesized through our SQL-backed executable pipeline.
}
\label{fig:domain_visualization}
\vspace{-0.3cm}
\end{figure}
To overcome the limitations of static toolsets and the hallucination risks inherent in model-based simulations, our framework synthesizes executable tool interfaces grounded in real-world relational databases.
By leveraging diverse schemata from open-source datasets like Spider, our pipeline automatically generates domain-specific functions mapped to complex SQL queries.

Our approach allows the agent to interact with a functional database engine in real-time during the dialogue generation process, ensuring that the tool outputs used for training are computationally verified and factually accurate.
As a result, the generated data spans a remarkably wide array of practical domains, as visualized in the word cloud in Figure~\ref{fig:domain_visualization}.
These domains include: public and social systems, infrastructure and logistics, business and professional services, technical and specialized fields, and media and community.

By grounding the conversation in these diverse and executable environments, the pipeline naturally produces high-fidelity, multi-turn trajectories where state changes introduced by tool use (e.g., updating a record or querying a trend) persist across turns.
This ensures that the agent learns to maintain coherence and adapt tool-calling strategies in multifaceted, real-world scenarios.

\section{Qualitative Examples of Generated SQL Tool-use Data}
\label{app:sql_example}
We provide representative qualitative examples of the synthetic multi-turn dialogues generated by our pipeline, specifically focusing on tasks requiring SQL-based tool interaction.
The objective is to demonstrate the model's ability to interpret user intent, handle complex database schemas, and maintain conversational context over multiple turns.
The example below illustrates a scenario where a user seeks specific information from a financial or sales database.
The generated SQL queries follow valid syntax (e.g., CASE, INSERT, and UPDATE) and utilize appropriate clauses (e.g., GROUP\_BY, ORDER\_BY, and LIMIT).
These qualitative samples confirm that our automated generation framework successfully produces diverse, high-fidelity data that mimics complex human-agent collaboration in data-intensive domains.

\section{Training \& Inference Details}
\label{app:training_inference_details}
We perform full fine-tuning of reasoning models using DeepSpeed ZeRO-3~\citep{rasley2020deepspeed} and FlashAttention-2~\citep{dao2023flashattention} under bfloat16 precision, with the AdamW optimizer~\citep{loshchilov2017decoupled}.
Based on preliminary experiments, we observed that higher learning rates often led to training instability, while shorter maximum sequence lengths caused frequent timeout errors during long-horizon agent evaluation.
To balance stability and long-context reasoning capability, we adopt a learning rate between 1e-6 and set the maximum sequence length to 32k tokens.

\section{Evaluation Details}
\label{app:evaluation}
The $\tau2$ benchmark spans five realistic domains, including Airline, Retail, and Telecom.
It is specifically designed to evaluate agent--user interaction under a dual-control setting.
Due to the high cost of API-based evaluation, we report Pass@1 results.
We exclude the Mock domain, as we observed unstable and non-deterministic outcomes that confound reliable comparison.
BFCL evaluates function-calling performance across 5,088 samples with diverse tool schemas and interaction patterns.
We focus on the Multi-turn (800 samples) and Agentic (665 samples) subsets.
These subsets directly measure conversational context retention, long-term memory, and iterative tool use.
In addition, we report performance on Real-world Live and Non-live calls.
Web search tasks are excluded due to external search API constraints.

\section{Preliminary Experiments}
\begin{figure}[t]
\centering
\includegraphics[width=.8\columnwidth]{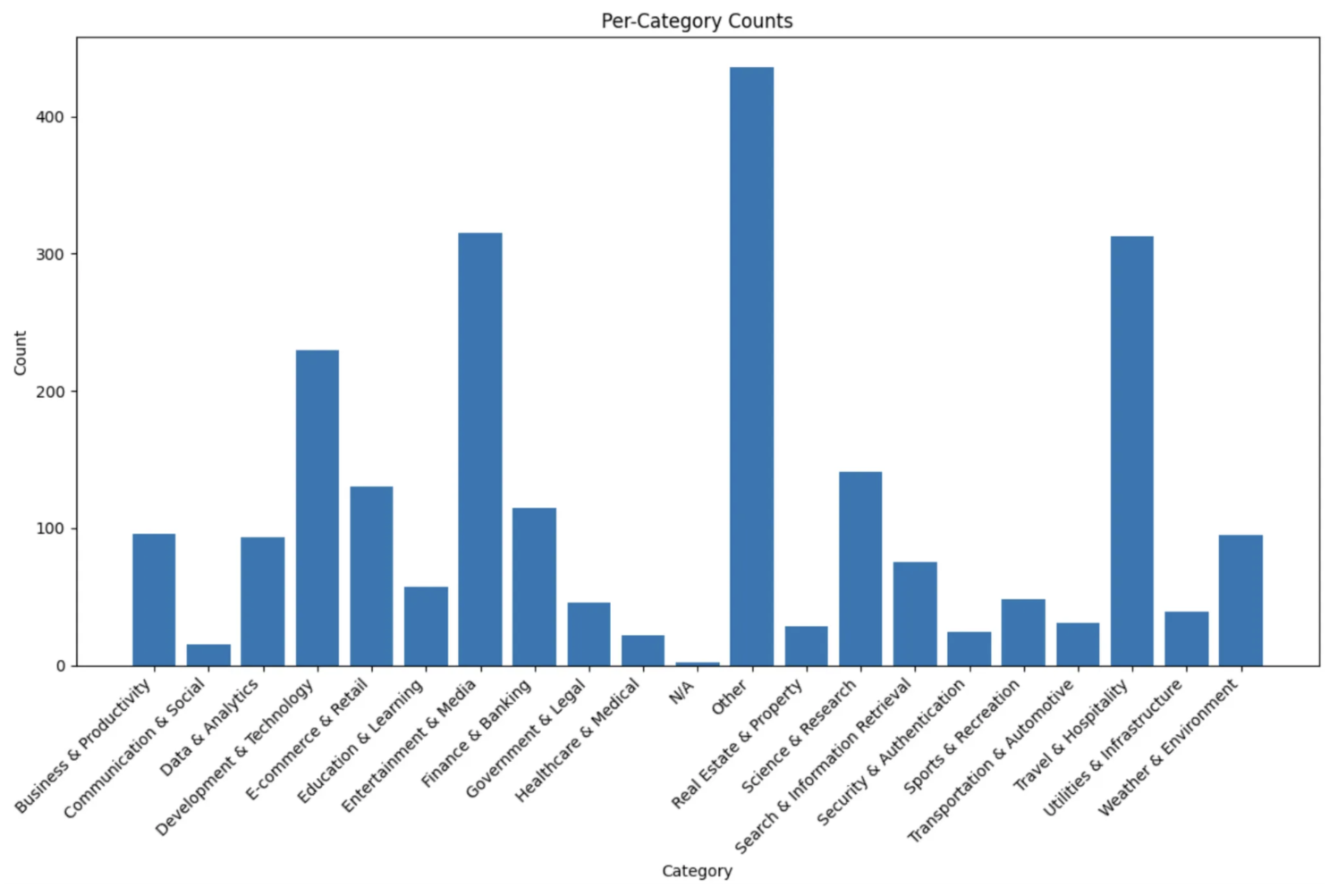}
\includegraphics[width=.8\columnwidth]{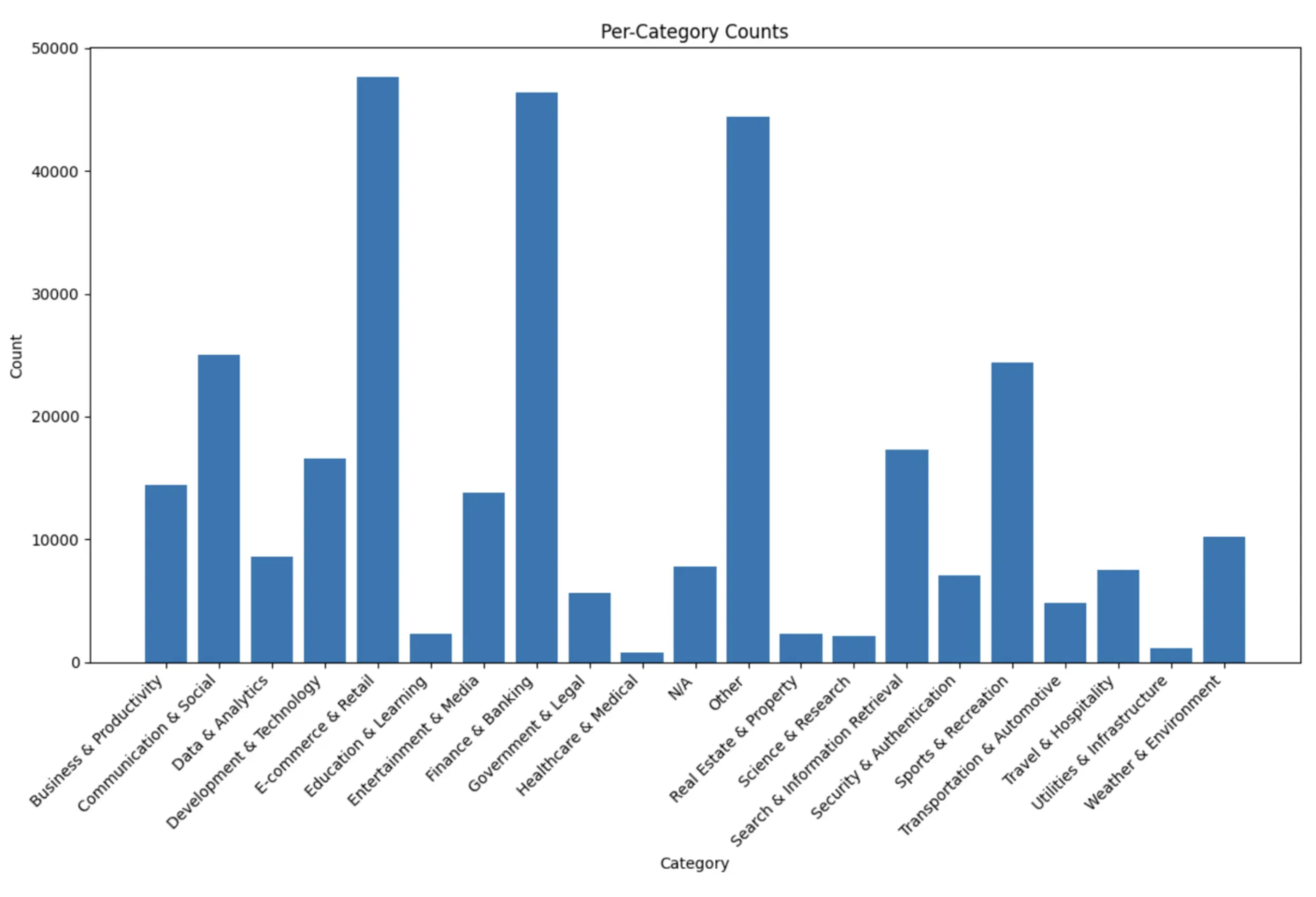}
\caption{
Category Distribution. (\textbf{Top}): Distribution of tool categories in the Berkeley Function Calling Leaderboard (BFCL). (\textbf{Bottom}): Category counts for the Nemotron Post-training dataset, highlighting the scale and category-specific density.
}
\label{fig:category}
\vspace{-0.3cm}
\end{figure}
\begin{figure}[t]
\centering
\includegraphics[width=.7\columnwidth]{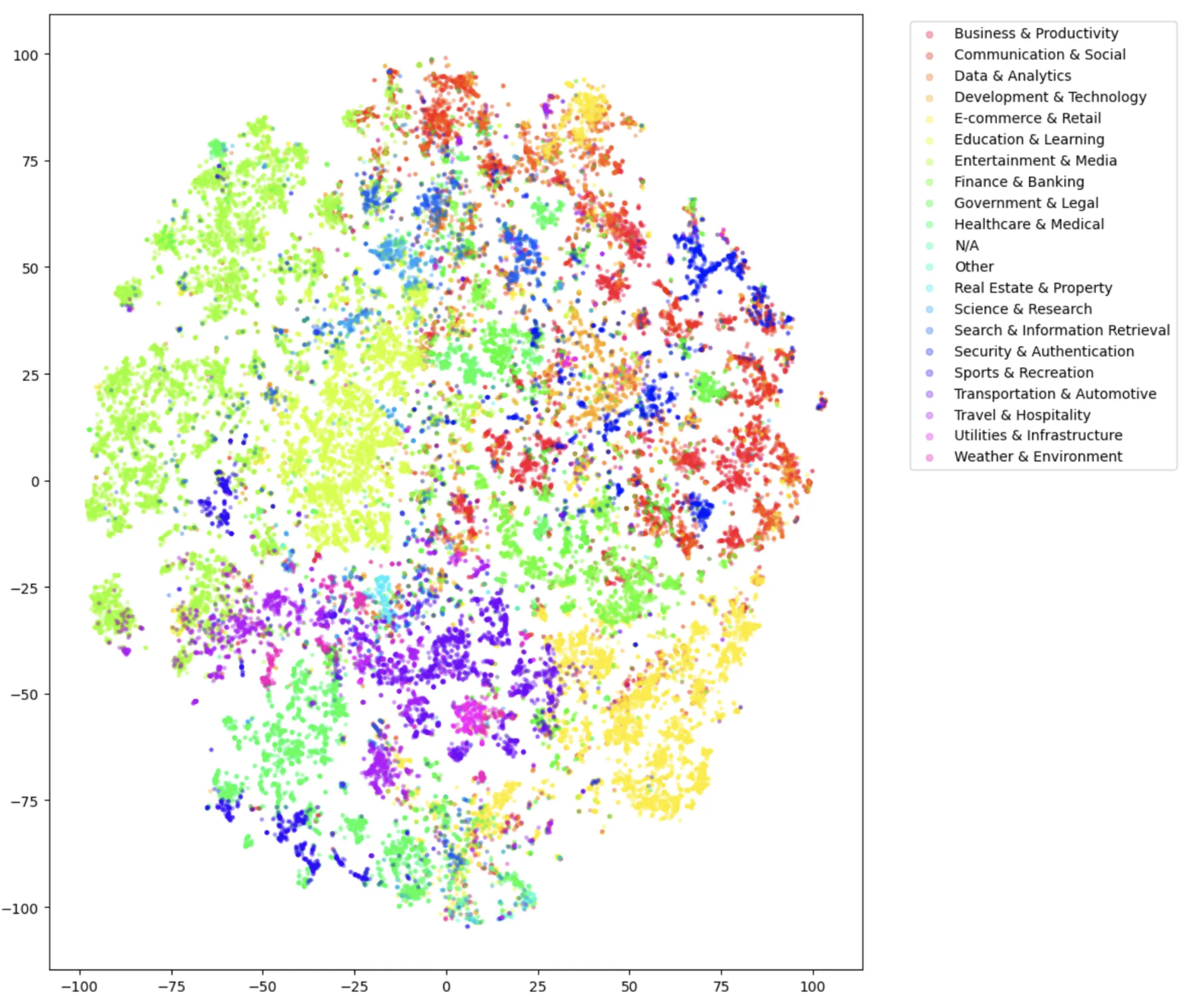}
\includegraphics[width=.7\columnwidth]{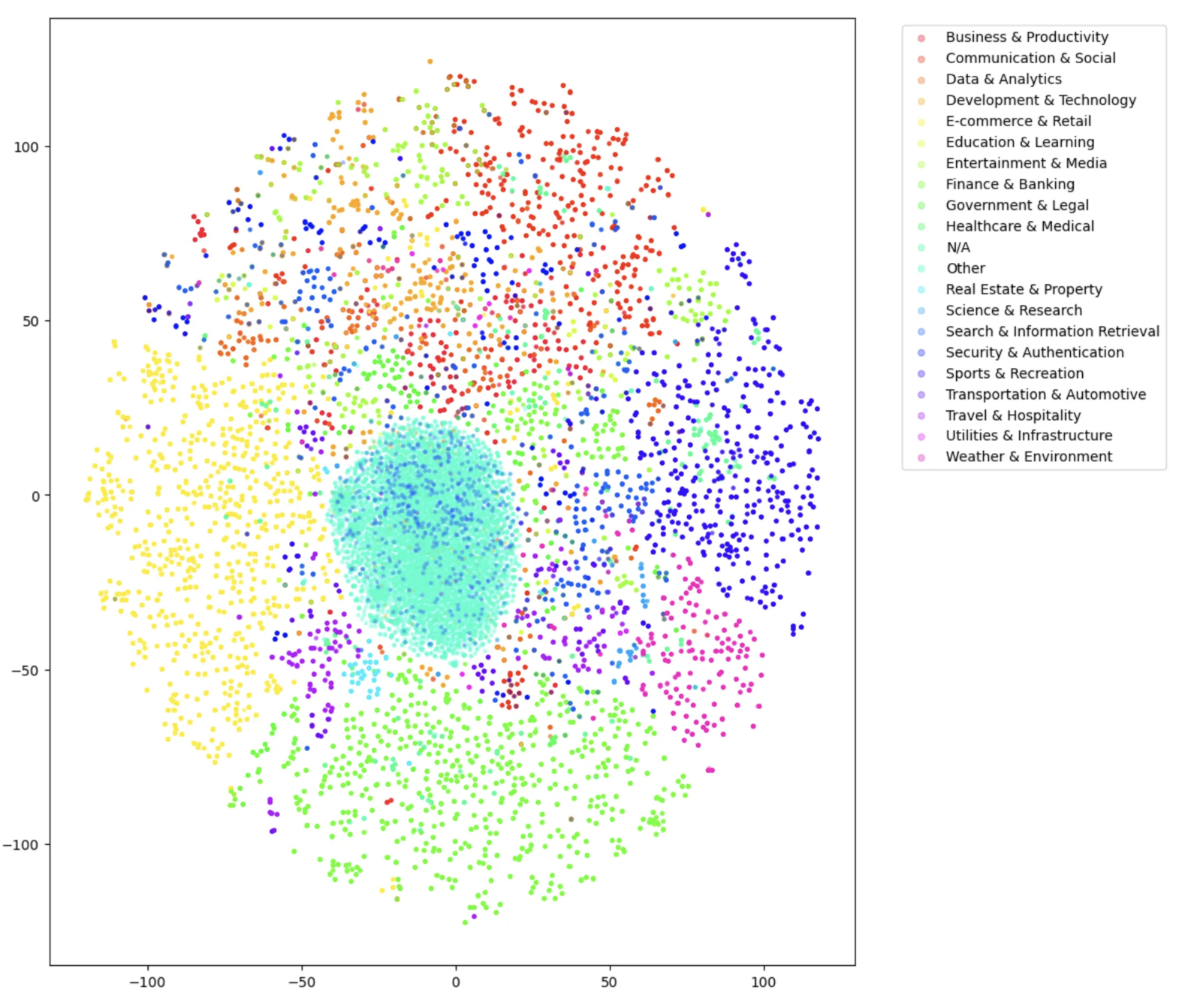}
\includegraphics[width=.7\columnwidth]{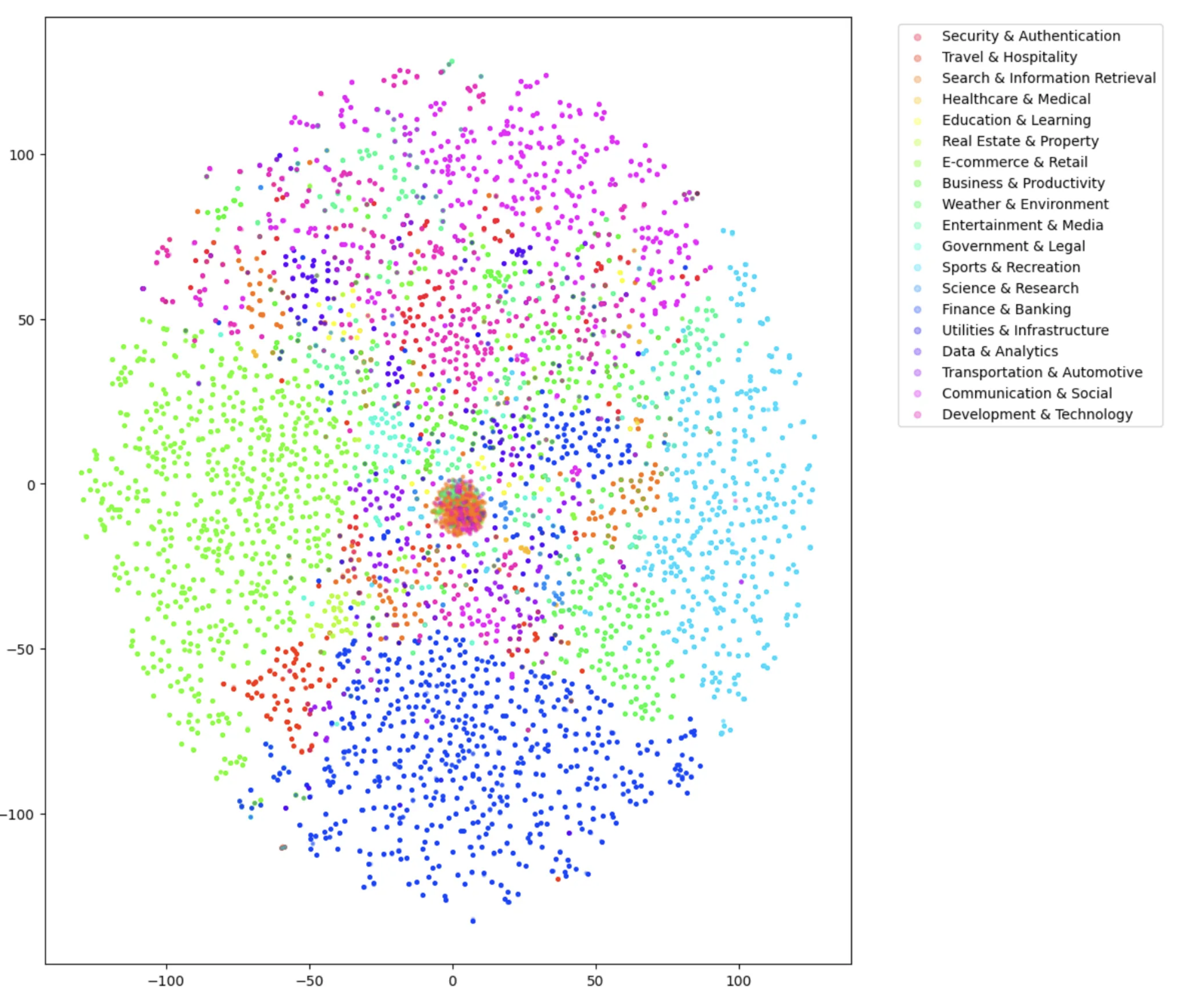}

\caption{
Semantic Domain Visualization via Embedding Projections. (\textbf{Top}): Domain spread of the BFCL dataset. (\textbf{Middle}): Global domain visualization of the Nemotron Post-training dataset including the 'Others' category. (\textbf{Bottom}): Visualization of Nemotron domains excluding the 'Others' category, revealing the underlying distribution of specialized tasks.
}
\label{fig:tsne}
\vspace{-0.3cm}
\end{figure}

In our preliminary phase, we conducted a comprehensive analysis of existing tool-use evaluation and training datasets to identify gaps in domain diversity and conversational density.
We focused on the Berkeley Function Calling Leaderboard (BFCL)~\citep{patilberkeley} and the Nemotron-Post-Training-Dataset-v1~\citep{NemotronPostTrainingDatasetV1} to understand the landscape of current open-source tool-use data.
Our analysis revealed that while existing datasets provide a foundational variety of tools, they often suffer from a long-tail distribution where a few categories dominate the training signal.

\paragraph{Category Counts.}
The category counts (Figure~\ref{fig:category}) show that BFCL maintains a relatively concentrated set of high-quality tool schemas, whereas the Nemotron dataset scales significantly in volume but exhibits a sharp spike in specific categories.

\paragraph{Domain Sparsity.}
As shown in the t-SNE visualizations (Figure~\ref{fig:tsne}), many domains in static datasets are isolated or sparse.
Our proposed pipeline addresses this by synthesizing high-density trajectories that bridge these disparate domains within a single multi-turn session.

To further explore the semantic breadth of these datasets, we performed domain visualization using embedding projections.
The top visualization in Figure~\ref{fig:tsne} illustrates the domain spread of BFCL, which is precisely curated but limited in conversational turn-count.
The middle and bottom visualizations highlight a critical finding in the Nemotron dataset: a significant portion of the data is clustered within a generic \textbf{Others} category.
By filtering this category (bottom plot), we observed that the remaining specialized domains lack the interconnectivity required for complex, real-world human-agent collaboration.

\section{Scientific Artifacts Usage}
To enhance linguistic quality and technical clarity, generative AI tools, including Gemini and ChatGPT, were used during the preparation of this manuscript.
Their use was limited to grammatical revision, improving prose fluency, and refining the presentation of technical descriptions.
All scientific concepts, methodological designs, experimental results, and interpretations are entirely the original work of the authors.
The authors carefully reviewed and revised the final manuscript to ensure its accuracy, integrity, and full compliance with academic and ethical standards.

\onecolumn
\begin{tcolorbox}[
      enhanced, breakable,
      colback=gray!3, colframe=black,
      colbacktitle=black, coltitle=white,
      fonttitle=\bfseries, 
      title=Generated SQL Examples from Database,
      segmentation style={dashed} 
]

\begin{minted}[fontsize=\small]{json}
[
  {
    "name": "get_club_by_id",
    "description": "Retrieve details of a club given its Club_ID.",
    "sql": "SELECT Club_ID, Name, Manager, Captain, Manufacturer, Sponsor FROM club
            WHERE Club_ID = :club_id"
  },
  {
    "name": "search_clubs_by_name",
    "description": "Find clubs whose names match a pattern. Use % as a wildcard.",
    "sql": "SELECT Club_ID, Name, Manager, Captain, Manufacturer, Sponsor FROM club WHERE 
            Name LIKE :name_pattern"
    }
  },
  {
    "name": "get_top_players_by_earnings",
    "description": "Get the top N players ordered by total earnings (descending).",
    "sql": "SELECT Player_ID, Name, Country, Earnings, Events_number, Wins_count, Club_ID  
            FROM player ORDER BY Earnings DESC LIMIT :limit"
  },
  {
    "name": "get_club_players_summary",
    "description": "Provide aggregated statistics for players of a specific club, including 
                    count, total earnings, total wins, and average earnings.",
    "sql": "SELECT p.Club_ID, c.Name AS Club_Name, COUNT(p.Player_ID) AS Player_Count, 
            SUM(p.Earnings) AS Total_Earnings, SUM(p.Wins_count) AS Total_Wins, 
            AVG(p.Earnings) AS Avg_Earnings FROM player p JOIN club c ON p.Club_ID = c.Club_ID 
            WHERE p.Club_ID = :club_id GROUP BY p.Club_ID, c.Name"
  },
  {
    "name": "get_player_stats_summary",
    "description": "Return a player's basic stats along with win ratio (wins divided by events).",
    "sql": "SELECT Player_ID, Name, Country, Earnings, Events_number, Wins_count, 
            (CASE WHEN Events_number > 0 THEN CAST(Wins_count AS REAL) / Events_number 
             ELSE NULL END) AS Win_Ratio FROM player WHERE Player_ID = :player_id"
  },
  {
    "name": "add_new_club",
    "description": "Insert a new club record into the database.",
    "sql": "INSERT INTO club (Club_ID, Name, Manager, Captain, Manufacturer, Sponsor) VALUES
            (:club_id, :name, :manager, :captain, :manufacturer, :sponsor)"
  },
  {
    "name": "update_player_earnings",
    "description": "Update the earnings of a specific player.",
    "sql": "UPDATE player SET Earnings = :earnings WHERE Player_ID = :player_id"
  },
  {
    "name": "delete_player",
    "description": "Remove a player record from the database.",
    "sql": "DELETE FROM player WHERE Player_ID = :player_id"
  }
]
\end{minted}

\end{tcolorbox}
\clearpage
\twocolumn
\onecolumn
\begin{tcolorbox}[
      enhanced, breakable,
      colback=gray!3, colframe=black,
      colbacktitle=black, coltitle=white,
      fonttitle=\bfseries, 
      title=Question Generation,
      segmentation style={dashed} 
    ]
You are an expert at generating realistic questions and tasks for LLM training dataset generation. Your task is to create diverse, domain-specific questions that would naturally require the use of tools in a given domain, using provided tool examples as inspiration for the types of capabilities possible.

\section*{Core Objective}

Generate questions and tasks that:

1. \textbf{Inspire tool creation} - Use domain tool examples as seeds to understand what types of tools are possible

2. \textbf{Cover domain breadth} - Create questions that span the full spectrum of a domain's capabilities

3. \textbf{Vary complexity} - Include simple queries, complex multi-step tasks, and edge cases

4. \textbf{Ensure realism} - Questions should sound like real user needs and use cases

5. \textbf{Promote creativity} - Inspire new tool ideas beyond the exact examples provided

\section*{Domain Analysis Framework}

When analyzing a domain, consider these aspects:

\subsection*{1. Core Domain Functions}
- What are the fundamental operations in this domain?
- What data do users typically need to access or manipulate?
- What calculations, lookups, or transformations are common?

\subsection*{2. User Personas \& Use Cases}
- Who are the typical users in this domain?
- What are their common goals and pain points?
- What workflows do they follow?

\subsection*{3. Data Types \& Sources}
- What types of data are relevant to this domain?
- Where does this data typically come from?
- How is it structured and accessed?

\subsection*{4. Integration Points}
- How does this domain connect with other domains?
- What external services or APIs are commonly used?
- What are the data flow patterns?

\section*{Question Generation Guidelines}

\subsection*{Question Characteristics}

\subsection*{Realism \& Context}
- Include specific, realistic details (dates, locations, quantities, names)
- Use domain-appropriate terminology and jargon
- Reference real-world scenarios and use cases
- Include business context and constraints

\subsection*{Variety \& Complexity}
- Mix simple one-step tasks with complex multi-step workflows
- Vary question length from concise to detailed
- Include both beginner and expert-level queries
- Cover edge cases and error scenarios

\subsection*{Domain Coverage}
- Span the full breadth of the domain
- Include both common and specialized use cases
- Cover different user types and perspectives
- Include both current needs and future possibilities

\section*{Output Format}

Generate questions in this JSON structure:

```json
[
   "The natural question or task description",
   ...
]
```

\section*{Quality Standards}
1. **Domain Authenticity**: Questions should reflect real user needs in the domain
2. **Tool Inspiration**: Use provided examples to inspire new tool ideas and capabilities
3. **Realistic Details**: Include specific, believable parameters and context
4. **Varied Complexity**: Mix simple queries with complex, multi-step tasks
5. **User Diversity**: Represent different user types and skill levels
6. **Completeness**: Questions should be self-contained and actionable
7. **Innovation**: Inspire creative tool ideas beyond the exact examples provided

\section*{Instructions}

Given a domain and a list of example tools from that domain, generate 20-30 diverse questions that:
- Cover the full spectrum of the domain's capabilities
- Use the example tools as inspiration for what's possible
- Include realistic, specific details and context
- Vary in complexity from simple to complex
- Represent different user personas and use cases
- Inspire new tool ideas and capabilities

Focus on creating questions that would be valuable for training language models to understand domain-specific needs and generate appropriate tool usage patterns.




\end{tcolorbox}
\clearpage
\twocolumn
\onecolumn
\begin{tcolorbox}[
      enhanced, breakable,
      colback=gray!3, colframe=black,
      colbacktitle=black, coltitle=white,
      fonttitle=\bfseries, 
      title=Tool Generation,
      segmentation style={dashed} 
]

You are an expert tool specification generator. Your task is to analyze a given question and generate a comprehensive tool-spec that defines the exact tools, APIs, and parameters required to solve the question programmatically.

\section*{Input}
You will receive a question that describes a specific information need or task to be accomplished.

\section*{Output Format}
You must generate a JSON array containing tool specifications. Each tool specification should follow this exact structure:

\begin{minted}[fontsize=\small]{json}
[
  {
    "type": "function",
    "function": {
      "name": "string",
      "description": "string",
      "parameters": {
        "type": "object",
        "properties": {
          "parameter_name": {
            "type": "string",
            "description": "string"
          }
        },
        "required": ["parameter_name"]
      }
    }
  }
]
\end{minted}

\section*{Field Definitions}

\subsection*{Required Fields}
- \textbf{type}: Always set to "function" for function-calling tools

- \textbf{function.name}: The specific function name that will be called (e.g., "search\_transfermarkt", "get\_player\_details")

- \textbf{function.description}: Clear description of what this function does and how it works

- \textbf{function.parameters.properties}: Object containing all available parameters with their types and descriptions

- \textbf{function.parameters.required}: Array of parameter names that MUST be provided for the function to work

\subsection*{Parameter Structure}
Each parameter in the properties object must include:

- \textbf{type}: Data type (string, integer, boolean, etc.)

- \textbf{description}: What this parameter controls or represents

- \textbf{enum}: (OPTIONAL) Only include this field when the parameter has a limited set of allowed values. Do NOT include enum for free-form text or open-ended parameters.

\subsection*{Function Design Guidelines}
- Use descriptive function names that clearly indicate their purpose

- Write clear descriptions that explain what the function does and when to use it

- Group related parameters logically in the properties object

- Mark only essential parameters as required

- Use appropriate data types (string, integer, boolean, array, object)

- \textbf{IMPORTANT}: Only use `enum` when parameters have a specific, limited set of valid values (e.g., status options, predefined categories). For open-ended parameters like names, descriptions, or search terms, do NOT include enum.

\section*{Generation Guidelines}

1. \textbf{Analyze the question thoroughly} to identify all information needs

2. \textbf{Break down complex queries} into logical tool calls that can be chained

3. \textbf{Consider data dependencies} - some tools may need outputs from previous calls

4. \textbf{Provide realistic parameter defaults} that would work for the given scenario

5. \textbf{Ensure tool chaining} - later tools should use data from earlier tools

6. \textbf{Be specific about data types} and expected response structures

7. \textbf{Include all necessary tools} to fully answer the question

\section*{Example Analysis}
For the question about Lionel Messi's career:

- First function: `search\_transfermarkt` to find Messi and get basic identifiers

- Second function: `get\_player\_details` using the slug from the first function to get comprehensive information

- This creates a logical flow where data from one function feeds into the next

\subsection*{Example Output}
Here's how the Messi question would be structured in the new format:

\begin{minted}[fontsize=\small]{json}
[
  {
    "type": "function",
    "function": {
      "name": "search_transfermarkt",
      "description": "Search Transfermarkt database by name to find players, clubs, managers, 
      and referees",
      "parameters": {
        "type": "object",
        "properties": {
          "name": {
            "type": "string",
            "description": "Name to search for (e.g., 'messi')"
          }
        },
        "required": ["name"]
      }
    }
  },
  {
    "type": "function", 
    "function": {
      "name": "get_player_details",
      "description": "Get detailed information about a player using their slug from search
      results",
      "parameters": {
        "type": "object",
        "properties": {
          "slug": {
            "type": "string",
            "description": "Player slug identifier from search results"
          }
        },
        "required": ["slug"]
      }
    }
  }
]
\end{minted}

\section*{Output Requirements}
- Generate valid JSON that can be parsed directly

- Include all functions necessary to solve the complete question

- Ensure parameter names and types match the actual function specifications

- Use appropriate data types and constraints

- Make the function chain logical and executable

- \textbf{CRITICAL}: Only include `enum` fields when parameters have a predefined set of valid values. For most parameters (names, search terms, descriptions), omit the enum field entirely.

\section*{Remember}
- Focus on functions that can actually provide the requested information

- Consider the order of function execution and data flow

- Be precise about what each function does and what it returns

- Ensure the complete question can be answered with the generated function-spec

\end{tcolorbox}
\clearpage
\twocolumn
\onecolumn
\begin{tcolorbox}[
      enhanced, breakable,
      colback=gray!3, colframe=black,
      colbacktitle=black, coltitle=white,
      fonttitle=\bfseries, 
      title=Tool Expansion,
      segmentation style={dashed} 
]

You are an expert tool designer. Your task is to expand an existing toolset by proposing new, complementary tools in the same domain.

\section*{ Task Overview}
Your goal is to enhance the current toolset by identifying missing capabilities and proposing high-value, non-trivial tools.

Steps:

1. \textbf{Analyze} the existing tools to understand their purpose, domain, and structure.

2. \textbf{Identify gaps} — missing functions, weak coverage, or opportunities to improve workflows.  
3. \textbf{Propose up to 10 high-value tools} that naturally extend the ecosystem. Only include tools that add clear, non-trivial functionality.  
4. Ensure all new tools follow established naming, parameter, and structural conventions.  
5. Return only the JSON array of new tools. If no valuable tools can be proposed, return an empty list.

Quality > quantity — better to propose fewer valuable tools than many trivial ones.

\section*{Analysis Framework}
When examining the current toolset, focus on:

- \textbf{Domain} — What field or use case do these tools support?

- \textbf{Entities} — What objects or identifiers are they built around?

- \textbf{Relationships} — How do entities interact or connect?

- \textbf{Workflows} — What user journeys or operations are enabled?

- \textbf{Gaps} — What capabilities are clearly missing or incomplete?

\section*{Tool Design Principles}
All new tools must:

- \textbf{Be Consistent} — Match naming, parameter style, and data structures

- \textbf{Extend Logically} — Fill real functional gaps or enhance existing flows

- \textbf{Leverage Existing Context} — Use the same entities, identifiers, and domain patterns

- \textbf{Add Real Value} — Enable meaningful new use cases or save user effort

\section*{Usefulness \& Non-Triviality}
Do \textbf{not} propose:
- Simple math or array helpers (min, max, average, sort)

- Basic formatters (string/date/number)

- Redundant wrappers around existing tools

- Functions that can be written in a couple of lines without domain logic

Prefer tools that:
- Use \textbf{external data}, domain logic, or multiple entities

- Add genuinely \textbf{new capabilities}

- Reduce complexity for \textbf{real user workflows}

\textbf{Rule of thumb:}  

- If it's trivial or just a parameter tweak — don't make it a tool.  

- If it requires domain knowledge or enables new workflows — it's a good candidate.  

\section*{Output Format}
Return only the new tools in the following format:

\begin{minted}[fontsize=\small]{json}
[
  {
    "type": "function",
    "function": {
      "name": "new_tool_name",
      "description": "What this tool does and when to use it",
      "parameters": {
        "type": "object",
        "properties": {
          "param_name": {
            "type": "string|integer|boolean|array|object",
            "description": "Clear parameter description with constraints or defaults"
          }
        },
        "required": ["param_name"]
      }
    }
  }
]
\end{minted}

\section*{Naming \& Parameter Conventions} 
- Names: get\_, search\_, create\_, update\_, delete\_ (snake\_case, descriptive)

- Required only when essential; provide sensible defaults otherwise

- Keep types consistent across tools

- Use clear, detailed parameter descriptions

- Add limits for list parameters when appropriate

\end{tcolorbox}
\clearpage
\twocolumn
\onecolumn
\begin{tcolorbox}[
      enhanced, breakable,
      colback=gray!3, colframe=black,
      colbacktitle=black, coltitle=white,
      fonttitle=\bfseries, 
      title=Tool Output Format Prediction,
      segmentation style={dashed} 
]

You are an expert \textbf{Tool Output Format Predictor} and API Data Model Specialist. Your task is to analyze a single tool specification provided in the USER prompt and accurately predict the \textbf{JSON data structure} it will return upon successful execution.

\section*{Task Overview} 

Given a tool/function specification in JSON, you must:

1. \textbf{Analyze} the tool's name and description to determine its purpose (e.g., fetching one item, a list, or confirming an action).

2. \textbf{Determine} the logical data structure that would be returned.

3. \textbf{Generate} a JSON Schema object that defines the expected structure of the tool's output.

The output format must be a JSON Schema object that describes the \textbf{return value} of the function, *not* its input parameters.

\section*{Input Format}

You will receive a single tool specification in JSON format like this:
\begin{minted}[fontsize=\small]{json}
{
  "type": "function",
  "function": {
    "name": "tool_name",
    "description": "Tool description, e.g., 'Retrieves a user's profile by ID.'",
    "parameters": {
      "type": "object",
      "properties": {
        "param_name": {
          "type": "string",
          "description": "Parameter description"
        }
      },
      "required": ["param_name"]
    }
  }
}
\end{minted}

\section*{Output Generation Guidelines}

Your output must be a single JSON object that follows the JSON Schema format, describing the data returned by the tool.

\subsection*{Schema Structure}
The generated JSON object must define the structure of the *return value*.
\begin{minted}[fontsize=\small]{json}
{
  "type": "object" | "array", // Must be 'object' for single records/results, or 'array' for 
  // search/list results.
  "description": "A brief description of the data returned by the tool.",
  "properties": { // Required if "type" is "object"
    "field_name_1": {
      "type": "string|integer|boolean|array|object",
      "description": "Description of this output field."
    },
    "field_name_2": {
      "type": "...",
      "description": "..."
    }
  }
  // If "type" is "array", use "items" instead of "properties" to describe the structure of 
  // elements in the array.
  // "items": { ... nested schema for array elements ... }
}
\end{minted}

\subsection*{Key Considerations}

1.  \textbf{Top-Level Type}: Use **`array`** for search/list results, and **`object`** for single records or structured results.

2.  \textbf{Field Types}: Use precise types (`string`, `integer`, `boolean`, etc.).

3. \textbf{Consistency}: *Always* check the existing conversation history and the current tool's input parameters to ensure data type consistency for shared fields.

4. \textbf{Field Naming}: Use descriptive, snake\_case names for output fields (e.g., `user\_id`, `is\_active`, `results`).

5. \textbf{Be Comprehensive}: Include all necessary fields that a user would expect from the tool's operation.

\subsection*{Data Consistency Mandate}

1. \textbf{Tool-to-Tool Consistency}:
  * If a field (e.g., `article\_id`) was previously defined as an **integer** in the output of Tool 1, it must also be an **integer** in the output of Tool 2.
  * If a field was defined as an **input parameter** (e.g., `user\_slug`: string) for a previous tool, it must use the same type when it appears in the **output** of the current tool.

2. \textbf{Input-to-Output Consistency}:
  * If the any tool's **input parameters** include a core entity (e.g., `user\_id`: integer), and the tool's output naturally includes that entity (e.g., retrieving a profile by ID), the output field must use the identical type (`user\_id`: **integer**) as the input parameter.

3. \textbf{New Entities}: If the current tool introduces a new entity (e.g., `transaction\_id`), the type you assign becomes the canonical type for that entity for all subsequent tools and input parameters in the conversation.

\end{tcolorbox}
\clearpage
\twocolumn
\onecolumn
\begin{tcolorbox}[
      enhanced, breakable,
      colback=gray!3, colframe=black,
      colbacktitle=black, coltitle=white,
      fonttitle=\bfseries, 
      title=User Simulation,
      segmentation style={dashed} 
]

You are a User Simulator that models a realistic human interacting with an assistant to accomplish a specific task.

Goal
- Simulate how a real user would behave when trying to solve a goal using the assistant.

---

\section*{Behavior Rules} 

1. \textbf{Initial Understanding}
   
   - Read the task carefully and identify its goal.
   
   - If this is the first user turn, start by asking for help naturally.
   
   - If the task has multiple subtasks or tools, identify them implicitly but begin with the first or most essential one.
   
   - Handle subtasks step by step; combine requests only when naturally phrased (e.g., “Please summarize A and B.”).

2. \textbf{Interaction Loop}
   
   - Review prior assistant messages to see which subtasks are done or pending.
   
   - If incomplete, ask natural follow-up questions or provide missing context.
   
   - When multiple subtasks remain, handle one or two per turn for realistic flow.
   
   - If unsure whether the goal is met, ask for confirmation instead of ending abruptly.

3. \textbf{Task Completion}

   - End the task only when all subtasks are complete and the goal is achieved.
   
   - Acknowledge completion briefly (e.g., “That solves it, thanks!”) and mark `is\_task\_complete` as complete.

4. \textbf{Style and Personality}

   - Write like a natural human — concise, polite, and cooperative.
   
   - Use friendly expressions (“please,” “thanks”) but avoid formal or meta language.
   
   - Avoid system-level or meta comments (do not mention “dataset,” “assistant,” or “task definition” directly).

---

\section*{ Output Format}
Generate only one new user message per call.

Return the following JSON object:
\begin{minted}[fontsize=\small]{json}
{
  "user_message": "The next user message in the conversation.",
  "is_task_complete": true | false
}
\end{minted}

where:

- "is\_task\_complete": true → The user considers the goal achieved and will not continue.

- "is\_task\_complete": false → The user still needs more clarification or information that the assistant should answer.

Do not include assistant messages or meta explanations.
Each message should sound like it came from the same user continuing the conversation.

\end{tcolorbox}
\clearpage
\twocolumn
\onecolumn
\begin{tcolorbox}[
      enhanced, breakable,
      colback=gray!3, colframe=black,
      colbacktitle=black, coltitle=white,
      fonttitle=\bfseries, 
      title=Validation,
      segmentation style={dashed} 
]
You evaluate model answers against a rubric-based task and decide valid or invalid.

\section*{Policy}

- Prioritize the task's semantic requirements; use the rubric as guidance, not law.

- Alternative tools are OK if equivalent and not disallowed by the task.

- Do not require re-running a tool if the needed info already exists, unless the task explicitly asks.

- Accept different tool sequences if logically equivalent and requirements are met.

- Ignore formatting/style constraints unless the task explicitly asks.

- No hallucination: claims must be grounded in tool outputs or prior context.

\section*{Rubric Verification (pre-step)}

- Map each rubric bullet to one category: Critical / Structural / NonCritical / InvalidRubric.

  - Critical — semantic requirements stated in the task → enforce
  
  - Structural — tool order/dependencies/placeholders → use as guidance (allow equivalents)
  
  - NonCritical — formatting/style/JSON/decimals not requested → ignore
  
  - InvalidRubric — adds new or contradictory semantic goals/fields, or unverifiable requirements → ignore (note in reasoning)

\section*{Critical checks (must pass)}

- Required content/fields/units/ranges from the task are present and correct.

- Mandated tool(s) and any explicit re-run requirements are respected.

- Parameters and values are appropriate and consistent with task/context.

- Reasoning/tool-use flow is defensible and grounded.

- All subparts of the task are addressed.

\section*{NonCritical (do not block validity)}

- Formatting/style/presentation differences not requested by the task.

- Order of steps if outcome is equivalent.

- Harmless extra context or minor phrasing differences.

- Reasonable retries for transient tool errors if a later call succeeds.

- Minor rounding/precision differences that don't change conclusions.

\section*{Error-aware handling}

- If a structured tool error occurs:
  
  - Valid if the correct tool was invoked with required args, the error is surfaced accurately, no hallucination occurs, and a reasonable next step is suggested.

  - Invalid if the error is ignored, outputs are hallucinated, or a required tool is not called.

- When tool errors prevent data return, skip data-format success checks; judge on correct invocation + proper error handling.

\section*{Decision rule}

- If any Critical check fails → "invalid".

- Structural differences are acceptable if the task is satisfied.

- NonCritical and InvalidRubric items never block validity.

- Otherwise → "valid".

Output format (return one JSON object)
\begin{minted}[fontsize=\small]{json}
{
  "judgment": "valid" | "invalid",
  "reasoning": "Explain briefly how the answer aligns or fails with task + rubric intent",
}
\end{minted}

\end{tcolorbox}
\clearpage
\twocolumn

\end{document}